\documentclass[a4paper,fleqn]{cas-sc}

\usepackage[authoryear]{natbib}
\usepackage{graphicx}
\usepackage{float}
\usepackage{algorithm}
\usepackage{algpseudocode}
\usepackage{color}
\usepackage{setspace}
\usepackage[nomarkers,figuresonly]{endfloat}

\usepackage{booktabs}
\usepackage{adjustbox}
\usepackage{subfigure}

\def\tsc#1{\csdef{#1}{\textsc{\lowercase{#1}}\xspace}}
\tsc{WGM}
\tsc{QE}
\tsc{EP}
\tsc{PMS}
\tsc{BEC}
\tsc{DE}

\usepackage{xfrac}

\begin{document}
\let\WriteBookmarks\relax
\def\floatpagepagefraction{1}
\def\textpagefraction{.001}
\shorttitle{Remote sensing colour image semantic segmentation of large herbivorous mammals trails}
\shortauthors{J.F.~D\'iez-Pastor et al.}

\title [mode = title]{Remote sensing colour image semantic segmentation of trails created by large herbivorous mammals}

\author[inst1]{José~Francisco D\'iez-Pastor}

\author[inst1]{Francisco~Javier Gonz\'alez-Moya}

\author[inst1]{Pedro Latorre-Carmona}

\author[inst2]{Francisco~Javier P\'erez-Barbería}

\author[inst3]{Ludmila~I. Kuncheva}

\author[inst1]{Antonio Canepa-Oneto}

\author[inst1]{\'Alvar Arnaiz-Gonz\'alez}

\author[inst1]{César García-Osorio}

\address[inst1]{organization={Department of Computer Engineering, Universidad de Burgos},
            addressline={Avenida Cantabria s/n},
            city={Burgos},
            postcode={09006},
            country={Spain}}

\address[inst2]{organization={Biodiversity Research Institute,  Spanish Research Council, University of Oviedo, Principado de Asturias)},
            addressline={c/ Gonzalo Gutierrez Quiros s/n},
            city={Mieres, Oviedo},
            postcode={33600},
            country={Spain}}

\address[inst3]{organization={School of Computer Science and Electronic Engineering, Bangor university},
            addressline={Dean Street},
            city={Bangor},
            postcode={LL57 1UT},
            country={United Kingdom}}

\begin{abstract}
Identifying regions where biodiversity faces elevated risk is essential for guiding conservation efforts and ecosystem monitoring. Large terrestrial herbivores act as allogenic engineers: through their feeding and trampling, they reshape soils, vegetation, and animal communities. One of the most visible signs of intense herbivore presence is the formation of grazing trails—interconnected bare-soil pathways created by repeated trampling. In this work, we benchmarked five semantic segmentation architectures, each combined with fourteen different encoder backbones, to automatically pinpoint grazing trails in aerial photographs. Our goal was to highlight areas of concentrated herbivory, thereby providing actionable information for habitat management and conservation planning. Most model-encoder pairings successfully delineated the trail networks, although a few combinations slightly underestimated their full extent. Overall, the UNet framework paired with the MambaOut encoder achieved the highest accuracy in mapping these paths.
The pipeline we introduce can be integrated into tools that track the emergence and evolution of grazing trails over time, offering valuable support for land-use management and biodiversity protection programs. To our knowledge, this represents the first instance of achieving competitive image segmentation performance for the detection and precise outlining of large herbivore trail systems.

\end{abstract}


\maketitle

\printcredits

\doublespacing

\section{Introduction} \label{sec:intro}

Large terrestrial herbivores are keystone species in ecosystem functions, as their type of biological interactions, known as herbivory (plant defoliation, trampling, defecation, and urination), have profound trophic effects.
They also act as allogenic ecosystem engineers, by creating physical changes in biotic or abiotic materials~\citep{jones_organisms_1994}.
Herbivory maintains, modifies, and creates new habitats and niches at various spatial scales, promoting habitat heterogeneity and, consequently, biodiversity~\citep{mills_keystone-species_1993, gordon_ecology_2019, filazzola_effects_2020}.
Low and medium intensities of herbivory positively affect biodiversity at both taxonomic and trait levels. However, high-intensity herbivory reduces habitat heterogeneity, compromising the number of niches and leading to a decrease in biodiversity~\citep{wieren_impact_2008}.

One of the landscape features created by the activity of large mammals, mainly ungulates, known as ``grazing trails''~\citep{chemekova_hillslope_1975, baocheng_livestock_2016, Jin2022TrackNetwork}.
These are narrow trails shaped by continuous animal trampling in a dominant direction toward a resource or to reduce energy expenditure during motion on steep terrain~\citep{may_mammal_1981, robbins_wildlife_1993}.
Trails can constitute networks of very different shapes, from resembling the contour lines of a topographic map, when trails develop on a steep slope~\citep{baocheng_livestock_2016, Jin2022TrackNetwork}, to radial networks due to the attraction to a concentrated resource such as a savanna waterhole~\citep{WashingtonAllen2004}.

A trail consists of a depressed longitudinal surface, often bare soil, bordered by raised shoulders~\citep{goudie2013, stavi_positive_2021}. These microtopography changes significantly influence some ecological processes, including water distribution, drainage, soil moisture~\citep{stavi_grazing-induced_2008}, soil erosion (both positive and negative)~\citep{higgins_grazing-step_1982, watanabe_soil_1994}, as well as nutrients, litter, root biomass, and vegetation distribution~\citep{stavi_grazing-induced_2008, stavi_positive_2021, hiltbrunner_cattle_2012}.

Since trails result from intense and concentrated activity of livestock and wild ungulates, monitoring their location over time and space is crucial to identifying herbivory hotspots that may compromise biodiversity in grazed ecosystems.
Developing automatic tools can facilitate the mapping and analysis of these complex landscape features in aerial images, supporting habitat conservation and land management programs.


The automatic characterisation of trails has received limited attention, with a few examples relying on landscape features such as relief, abrupt elevation changes, topographic position, among others~\citep{Godone_et_al_2018, DIAZVARELA2014117, Sas2012DETECTIONOO, SOFIA2014123, Bailly_Lavavaseur_2012, PIJL2020101977}.

To the best of our knowledge, the only study that aimed to automatically detect livestock trails from high-resolution satellite imagery is the work by~\cite{hellman_detection_2020}.
However, their method is limited to predicting trail presence at the patch level ($37\times37$m).
We improve upon this approach by using semantic segmentation, enabling pixel-level detection that ultimately provides an effective means of reconstructing the full extent of grazing trails.

Semantic segmentation~\citep{Hao2023} is currently an active area of research that started within the computer vision community, but extended later to a wide and diverse group of application fields~\citep{Guo2018, Hao2020, Minaee2022}.
It is somewhat related to image classification since it produces per-pixel category prediction rather than of an image-level prediction.

Semantic segmentation using deep learning architectures has been applied for artificial structures, like roads and buildings, or even parts of cities, as shown by \cite{Jamali2024}, \cite{fibaek2024phileo}, \cite{Hong2023} and \cite{Wurm2019}.
This approach takes into account both local information provided by convolutional neural network methodologies and global contextual information provided by architectures like the so-called vision transformers~\citep{khan2022transformers}.

Other semantic segmentation strategies are applied to identify natural objects/structures, such as rivers~\citep{Chen2023, Wieland2023}, ice in rivers and open seas~\citep{Zhang2020, Zhang2022}, forests~\citep{Bragagnolo2021, BorgesDaCosta2022}, or agricultural structures~\citep{Jadhav2018, Luo2023}.

In some cases, the application areas may involve considering truly remote sensing images, while in others the images are
captured much closer to the objects of interest, which may not strictly qualify as remote sensing images~\citep{Dong2021, Hosseini2022}.

Another approach has been to consider the identification of natural or artificial structures as a classification-related problem~\citep{Pal2020,Parajuli2022}.
For instance, water bodies are identified in remote sensing images acquired by the European Space Agency (ESA) Sentinel-2 satellite~\citep{s2mission} by extracting patches and classifying their central pixels as either water or non-water pixels.


Note that any local-based approach would not consider non-local information that an expert might use to determine a grazing trail such as its shape, extent, width, etc.
This is the rationale for using a semantic segmentation approach, which not only identifies the pixels constituting a trail, but also incorporates non-local information during the segmentation process (\textit{i.e}., features from the image that are further away but still influence the correct identification or classification of the pixels forming a trail).

The aim of this research is to assess the effectiveness of different semantic segmentation methodologies for mapping grazing trails, ensuring that the methods enable the detection of these landscape features without relying on additional topographic information. This presents a challenge due to the wide variety of shapes, continuity, and complexity within these trail networks.



\section{Materials and methods} \label{sec:materials_methods}
This section details the image processing techniques, from image acquisition to the algorithms used for its processing.
The main features regarding the specific geographical area and morphology considered in our analysis, as well as the type of processed images are presented in subsection~\ref{subsec:studyarea}.
Then, the creation of the \textit{groundtruth} images is explained in Subsection~\ref{subsec:gt-gen}.
Semantic segmentation and the architectures used in the study are detailed in Subsection~\ref{subsec:seg}, and finally, in Subsection~\ref{subsec:exp-setup} the details of the experimental setup are presented.

\subsection{Aerial images} \label{subsec:studyarea}
\begin{figure}
    \centering
    \includegraphics[width=0.6\linewidth]{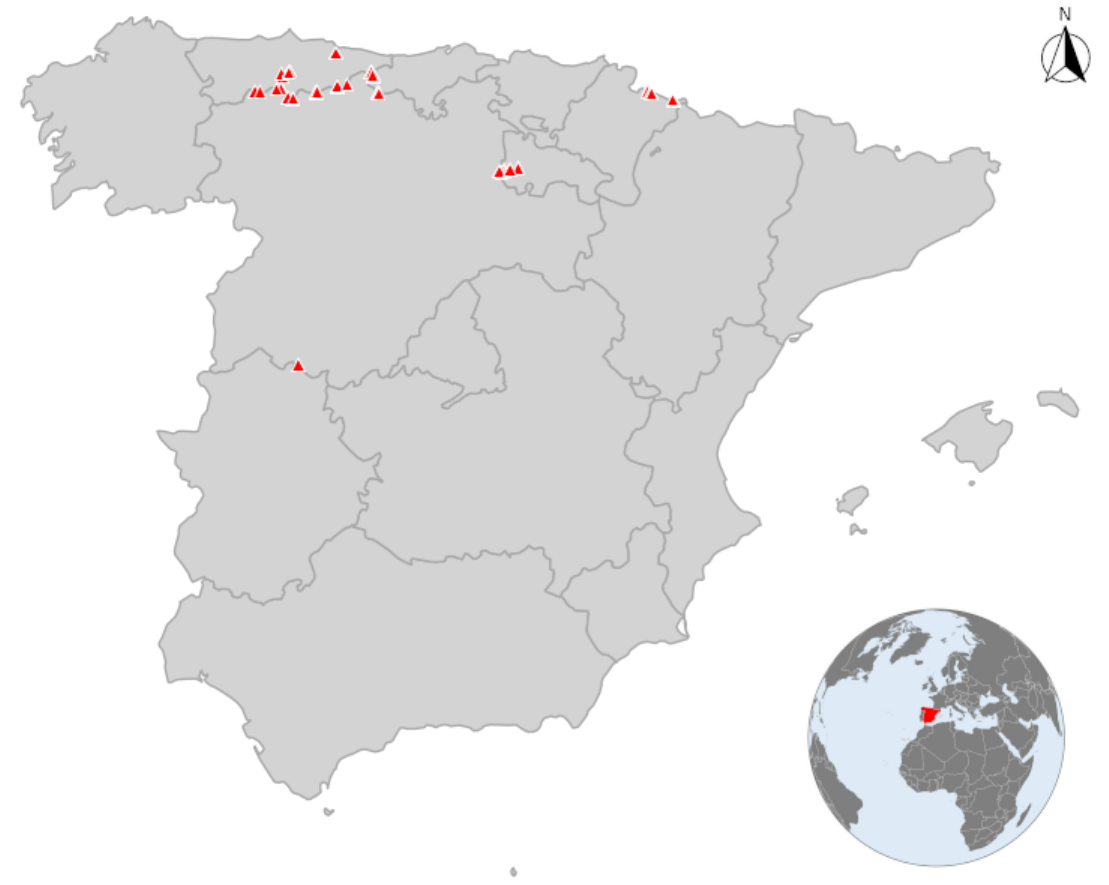}
\caption{Study area. Map of Spain showing the areas (red triangles) images were obtained from. The selected mountain systems correspond to: (i) ``Cantabrian'' mountain range, (ii) ``Palencia'' mountain range, (iii) ``Pyrenees'', (iv) ``Sierra de la Demanda'', and (v) ``Sierra de Béjar''.}
\label{fig:geo}
\end{figure}

One hundred (100) aerial images from grazed ecosystems in five mountain ranges in Spain: (1) ``Cantabrian'' mountain range; (2) ``Pyrenees''; (3) ``Sierra de la Demanda''; (4) ``Sierra de B\'ejar'', and (5) ``Montaña Palentina'', were used (Figure~\ref{fig:geo}).
According to data from regional administrations, all these areas show livestock activity and have been subject to both local and transhumant livestock movements since at least the Middle Ages.

\begin{figure}
    \centering
    \includegraphics[width=1\linewidth]{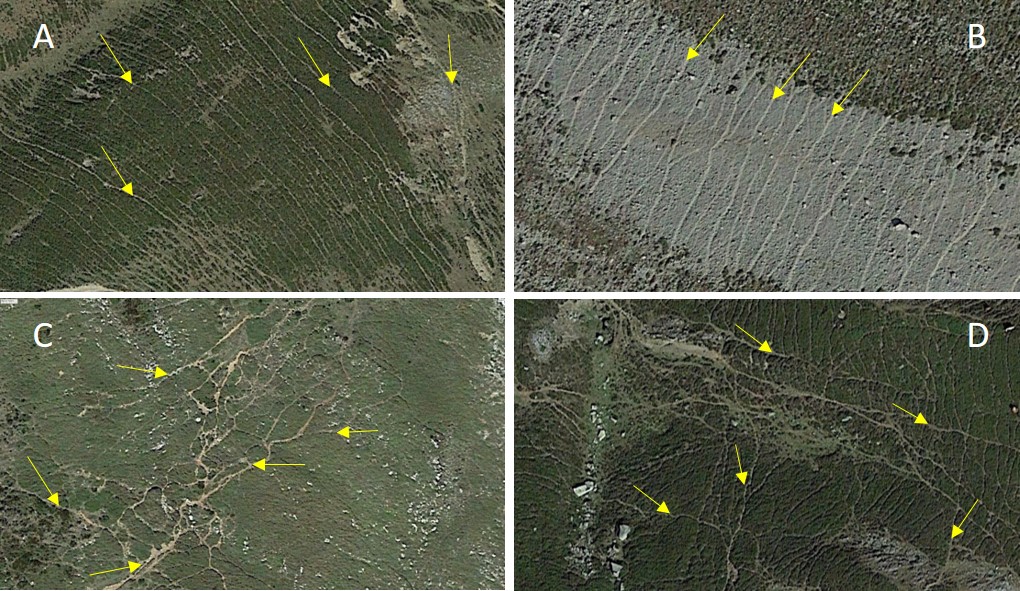}
    \caption{Examples of different grazing trails networks in the ``Cantabrian'' mountain range. (A) Parallel trails on heather, their layout dominant direction is perpendicular to the slope, 43° 1'37.14" N, 5°30'4.25" O WGS84; (B) As (A) but on a rocky slope, coordinates 43°11'17.56" N, 4°45'24.56" O; (C) Trails on mountain grassland, no dominant layout direction, coordinates 43° 2'25.66" N, 6°13'20.43" O; (D) Trails on heather, two dominant directions, coordinates 43° 1'35.50" N, 5°29'40.07" O. Some trails are marked with arrows.}
    \label{fig:Paths}
\end{figure}

One of the co-authors with extensive experience in grazing ecology, used orthoimages from Google Earth Pro to examine these areas and conducted visual transects on the images to identify grazing trails.
Then, in images projected at a height of 100 metres above the ground (approximate surface $70m \times 115m$), the observer selected $100$ images that represented the variety of shapes and densities of the trail networks in the area (Figure~\ref{fig:Paths}). The centre of each image was recorded as a ``.kmz'' file for georeferencing purposes, and the image was saved as a ``.jpg'' file at maximum resolution ($8192\times4968$ pixels). The selected trails ranged from an altitude (h) between $1091$ and $1880$ meters above sea level (where the mean height is: $\bar{h}= 1487$, and its standard deviation: $\textrm{std}(h) = 234.7$)\footnote{Supplementary information can be found in the  
Appendix A}.

\subsection{\textit{Groundtruth} image generation} \label{subsec:gt-gen}
To train the models and assess the segmentation quality, \textit{groundtruth} images were created by the same expert.
A copy of each image was imported to ImageJ~\citep{schneider_nih_2012} and all visible trails were drawn along the centreline using the yellow pencil tool (10 pixels in width).
This RGB-labelled image \( I_{\text{rgb}} \) was converted to a binary-labelled image \( C \), the \textit{groundtruth} \( M \) was obtained following the methodology described by~\cite{Mnih2010}.

Let \( I_{\text{rgb}} \) be an image in RGB format, manually labelled by the expert with yellow lines marking the trails.
Let \( I_{\text{hsi}} \) be the image resulting from converting \( I_{\text{rgb}} \) to the HSI colour space (being the three coordinates, normalised to the $[0, 1]$ interval).
Let \( P_{\text{ref}} \) be the reference pixel (colour) of the labels in the HSI space, given by $ P_{\text{ref}} = [H_{\text{ref}} = 0.6,\; S_{\text{ref}} = 1,\; I_{\text{ref}} = 1]$.

Since the pencil tool creates traces with some degree of smoothing, further processing is required to generate the corresponding binary masks. For each pixel \( P_{ij} = [H_{ij}, S_{ij}, I_{ij}] \) in \( I_{\text{hsi}} \), the Euclidean distance \( D_{ij} \) between \( P_{ij} \) and \( P_{\text{ref}} \) is obtained as:
\[
D_{ij} = \sqrt{\big(H_{ij} - H_{\text{ref}}\big)^2 + \big(S_{ij} - S_{\text{ref}}\big)^2 + \big(I_{ij} - I_{\text{ref}}\big)^2}
\]

These distances are then scaled by dividing them by the maximum distance $D_{\text{max}}$ in the entire image:
\[
D_{ij-{\text{norm}}} = \frac{D_{ij}}{D_{\text{max}}}
\]

Finally, a binary-labelled mask image \( C \) is created where pixels with \( D_{\text{norm}} < \textit{Th} \) are labelled as true, \textit{i.e.}:
\[
C_{ij} = \begin{cases}
\text{1}, & \text{if } D_{ij-{\text{norm}}} < \textit{Th} \\
\text{0}, & \text{otherwise}
\end{cases}
\]

In our case, we set $\textit{Th} = 3$. $C_{ij}$ is then $1$ if the $(i,j)$ pixel location in the remote sensing image $S$ belongs to a trail centreline, and 0 otherwise. The mask $C$ is later used to define the \textit{groundtruth} map $M$ as follows:

\begin{equation}
M_{ij} = \exp\left[-\frac{d_{ij}^{2}}{\sigma^{2}}\right]
\end{equation}

\noindent where $d_{ij}$ is the Euclidean distance between the location with coordinates $(i,j)$ in the image, and the nearest nonzero pixel in $C$.  $\sigma$ is a smoothing parameter that depends on the scale of the aerial images.

\cite{Mnih2010} set the parameter $\sigma$ such that the distance equivalent to $2 \sigma + 1$ pixels roughly corresponds to the width of a typical road lane.
In our case $\sigma$ was set to 16, a value that visually corresponds to half the average width of the paths in the images in the original training set. $M_{ij}$ can be interpreted as the probability that location $(i,j)$ belongs to a trail, given that it is $d_{ij}$ pixels away from the nearest centreline pixel.

The aerial images were scaled down by a factor of 8, in order to reduce the computational cost of the experiments (scale factors $2$ and $4$ were also tested without detecting any difference in the results).

The dataset consisting of RGB images and their corresponding \textit{groundtruth} images was generated using HuggingFace\footnote{HuggingFace is a repository of ready-to-use datasets for machine learning applications}, accessible on HuggingFace Hub\footnote{\url{https://huggingface.co/datasets/jfdpastor/Comp_Geo_Herb_Trails_8x}}.


 \subsection{Semantic segmentation} \label{subsec:seg}


In this paper, five semantic segmentation architectures to identify grazing trails in aerial images were used:

\begin{itemize}
    \item U-Net (UNet) is the \textit{seminal} work that proposed the so-called \textit{fully convolutional network} architecture/methodology for computer vision~\citep{Ronneberger2015}. On of its main advantages is hat it works with very few training images. It also yields precise segmentations. \textit{Fully convolutional networks} supplement a usual \textit{contracting network} by successive layers, where pooling operators are replaced by upsampling operators.
    In a \textit{contracting network}, the spatial information is reduced while the information about the features is increased.
    Therefore, upsampling operators make these layers increase the resolution of the output.
    High resolution features from the contracting path are combined with the upsampled output.
    A successive convolution layer can then learn to assemble a more precise output based on this information.
    \item Feature Pyramid Networks (FPNet) are generic feature extractors that exploit multi-level feature representations in an inherent and pyramidal-type hierarchy~\citep{Lin2017}.
    They use a top-down architecture with \textit{lateral} connections to fuse high-level semantic information into middle and low levels, with little extra cost.
    In particular, the aim in this method is to leverage a convolutional neural network's pyramidal feature hierarchy, which has semantics from low to high levels, and build a feature pyramid with high-level semantics throughout it.
    The resulting feature pyramid network is general and the focus is on sliding window strategies, and region-based detectors.
    FPNets are generalized to instance segmentation proposals.
    They take a single-scale image of an arbitrary size as input data, and give as an output a series of proportionally sized feature maps at multiple levels.
    This process is independent of the convolutional architectures for feature extraction.
    \item The Pyramid Scene Parsing Network (PSPNet) method tries to take advantage of the capability to use global context information by an aggregation methodology of different-region based features~\citep{zhao2017pspnet}.
    In a deep neural network, the size of the so-called \textit{receptive field} (i. e., the size of the area in the input image that creates the feature) may roughly indicate how much context information we use/consider. In order to further reduce context information loss between different sub-regions, researchers proposed a hierarchical global prior, containing information with different scales and varying among different sub-regions~\citep{zhao2017pspnet}.
    \item The transformer framework for semantic segmentation (Semantic Segmentation with Transformers or Segformer) can be understood as a semantic segmentation method which unifies the so-called \textit{transformers} with lightweight multilayer perceptron (MLP) decoders~\citep{Xie2021}. A vision transformer (ViT) decomposes an input image into a series of patches, serializes each patch into a vector, and maps it to a smaller dimension. These vector embeddings are then processed by a transformer encoder as if they were token embeddings. ViTs were designed as alternatives to convolutional neural networks.
    In particular, they present a new positional-encoding-free hierarchical transformer and a lightweight All-Multilayer Perceptron\footnote{Multilayer Perceptron is a classical artificial neural network that consists of fully connected artificial neurons which are organised in layers.
    It has at least three layers, input layer (the first one), output layer (the last one and gives the prediction), and hidden layer (placed in between the other two and can be more than one).} (MLP) decoder design.
    \item The Unified Perceptual Parsing Network (UperNet) is a network based on FPNet~\citep{Xiao2018}, previously described in~\citep{Lin2017}.
    Its main drawback is that, even when the theoretical receptive field of deep CNNs might be large enough, the empirical receptive field of deep CNNs may also be relatively smaller.
    To address this issue, a Pyramid Pooling Module (PPM) from PSPNet was proposed~\citep{zhao2017pspnet}.
    This PPM is applied on the last layer of the backbone network, just before feeding it into the top-down branch, in FPNet.
\end{itemize}

These methods were combined with 14~encoders, listed in Table~\ref{table:encoders} along with their number of parameters.
Encoders accomplished two main purposes: (a) to generate a high-dimensional feature vector from the input image, and (b) to aggregate features across multiple resolution levels.
Similarly, decoders also serve two key functions: (a) to produce a semantic segmentation mask from the high-dimensional feature vector, and (b) to decode the multi-level features aggregated by the encoder.

\begin{center}
\begin{table}
\begin{tabular}{@{}l r l@{}}
     \toprule
        \textbf{Encoder} & \textbf{\# of parameters}~(M) & \textbf{Reference} \\
     \midrule
         ConvNeXt (small) & 50 & \cite{liuconvnet2022} \\
         EfficientViT B3 & 12 & \cite{caiefficientvit2023} \\
         EfficientNet B7 & 63 & \cite{tanefficientnetrethinkingmodelscaling2020} \\
         MambaOut (Base) & 85 & \cite{Gumamba2024}\\
         Mobilenet v3 (Large) & 15 & 
         \cite{Howard2019} \\
         SAM 2 Hiera (Large) & 224 & \cite{ryalihiera2023} \\
         Densenet-161 & 26 & \cite{huangdenselyconnectedconvolutionalnetworks2018}\\
         Inception v4 & 41 & \cite{szegedyinceptionv4inceptionresnetimpactresidual2016} \\
         MIT B5 & 81 & \cite{Xie2021} \\
         MobileOne s4 & 12 & \cite{Vasumobileone2022}\\
         ResNet-34 & 21 & \cite{He2016}\\
         Xception-71 & 20 & \cite{cholletxceptiondeeplearningdepthwise2017}\\
         Vgg16 & 14 & \cite{simonyandeepconvolutionalnetworkslargescale2015} \\
         Vgg19 & 20 & \cite{simonyandeepconvolutionalnetworkslargescale2015} \\
     \bottomrule
\end{tabular}
\caption{List of encoders used: name, number of parameters (in millions), and reference. All of them were pretrained with the so-called \textit{ImageNet} dataset.}
\label{table:encoders}
\end{table}
\end{center}

\subsection{Experimental setup} \label{subsec:exp-setup}
To assess the best combination of architecture and encoder for grazing trail identification, the Pytorch Segmentation Model library was used~\citep{Iakubovskii2019}.
This framework allows different architectures to be paired with various encoders, facilitating a thorough evaluation of multiple configurations.

For the assessment of the performance of the models, a $10-$fold cross-validation strategy was followed on every architecture-encoder pair, \textit{i.e.,} $5 \times 14 = 70$ combinations.
In every single fold, from the set of $100$~ aerial images (with their corresponding 100~\textit{groundtruth} images), $80$ images were used for model training, $10$ images for internal model validation and, finally, $10$ images were used for testing.
The set of training, validation, and test images alternates on every fold to ensure the use of each instance in the test part only once.

To compare the performance of the different models, the following measures were used: the Intersection Over Union (IoU, also called the Jaccard coefficient), \textit{Precision}, \textit{Recall}, and the \textit{F1}-score. IoU can be used to measure the similarity between finite sample sets ($\mathbf{A}$ and $\mathbf{B}$), and it is defined as follows:

\begin{equation}
\text{IoU}(\mathbf{A},\mathbf{B}) \equiv J(\mathbf{A},\mathbf{B}) = \frac{|\mathbf{A}\cap\mathbf{B}|}{|\mathbf{A}\cup\mathbf{B}|}
\end{equation}

\noindent with: $0 \leq J(\mathbf{A},\mathbf{B}) \leq 1$. In our case, $\mathbf{A}$ and $\mathbf{B}$ would be the \textit{groundtruth} image, and the image segmented by the corresponding method, respectively. In addition, a pixel associated with a trail can be regarded as belonging to the positive class.
Taking into account the concept of \textit{true positives} $\textit{TP}$ (case where a pixel belongs to a trail and the segmentation method classifies it as trail), \textit{true negatives} $\textit{TN}$ (a pixel does not belong to a trail and the segmentation method classifies it as not belonging to a trail), \textit{false positives} $\textit{FP}$  (a pixel does not belong to a trail and the segmentation method classifies it as belonging to a trail), and \textit{false negatives} $\textit{FN}$ (a pixel belongs to a trail but the segmentation method classifies it as not belonging to a trail), the Precision, Recall, and F0 measures would be defined as follows:
$\text{Precision} = \frac{\textit{TP}}{\textit{TP}+\textit{FP}}$, $\text{Recall} = \frac{\textit{TP}}{\textit{TP}+\textit{FN}}$, $\text{F1} = 2\times \frac{\text{Precision} \times \text{Recall}}{\text{Precision} + \text{Recall}}$




To compare the performance of the different architectures and encoders, average rankings were used~\citep{wilcoxon1992individual}.
For each encoder, architectures were ranked based on their mean IoU scores, with tied architectures assigned the average of their competing ranks (\textit{e.g.}, two architectures tied for first place receive a ranking value of 1.5). The final performance metric for each architecture was derived by averaging these per-encoder rankings across all encoders, where lower average rankings indicate superior overall performance.

Bayesian tests~\citep{Benavoli2017} were used to assess whether the differences between algorithms are significant.
In particular, the Bayesian interpretation of the $t$-test was considered.
This approach is particularly suitable for results obtained when applying a cross-validation strategy.
This test takes as a parameter the width of the region of practical equivalence (ROPE), or in other words, the maximum difference that can exist between two results in order to be considered equivalent.
In this work, a region of practical equivalence of $0.01$ has been used.

The Bayesian test gives three probabilities as its outcome: (a) $p_{\text{left}}$, the probability that the first method outperforms the second, (b) $p_{\text{rope}}$, the probability that both methods can be considered \textit{equivalent}, and (c) $p_{\text{right}}$, the probability that the second method outperforms the first.
Whether one method is \textit{practically equivalent} to the other is determined by the ROPE parameter.

The test uses the results from a $10$-fold cross-validation and applies Monte Carlo simulation to generate $5.0\times10^{4}$ samples.
Based on these samples, a distribution is modelled, and the three probabilities are estimated.
Bayesian tests were performed using the \textit{Baycomp}\footnote{\url{https://baycomp.readthedocs.io/}} library.

\section{Results} \label{res}

\begin{figure*}
    \centering
    \subfigure{\includegraphics[width=.48\textwidth]{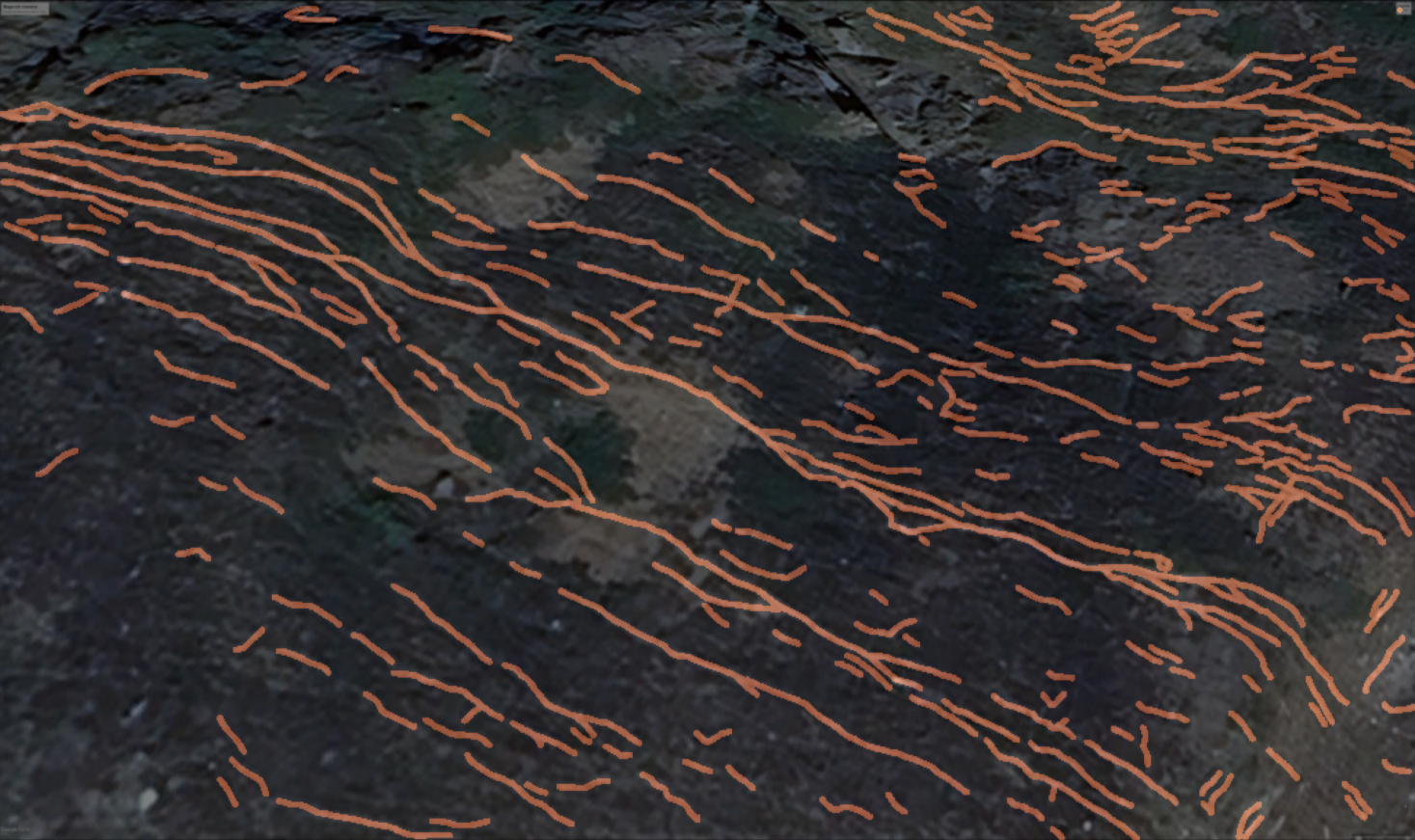}}~
    \subfigure{\includegraphics[width=.48\textwidth]{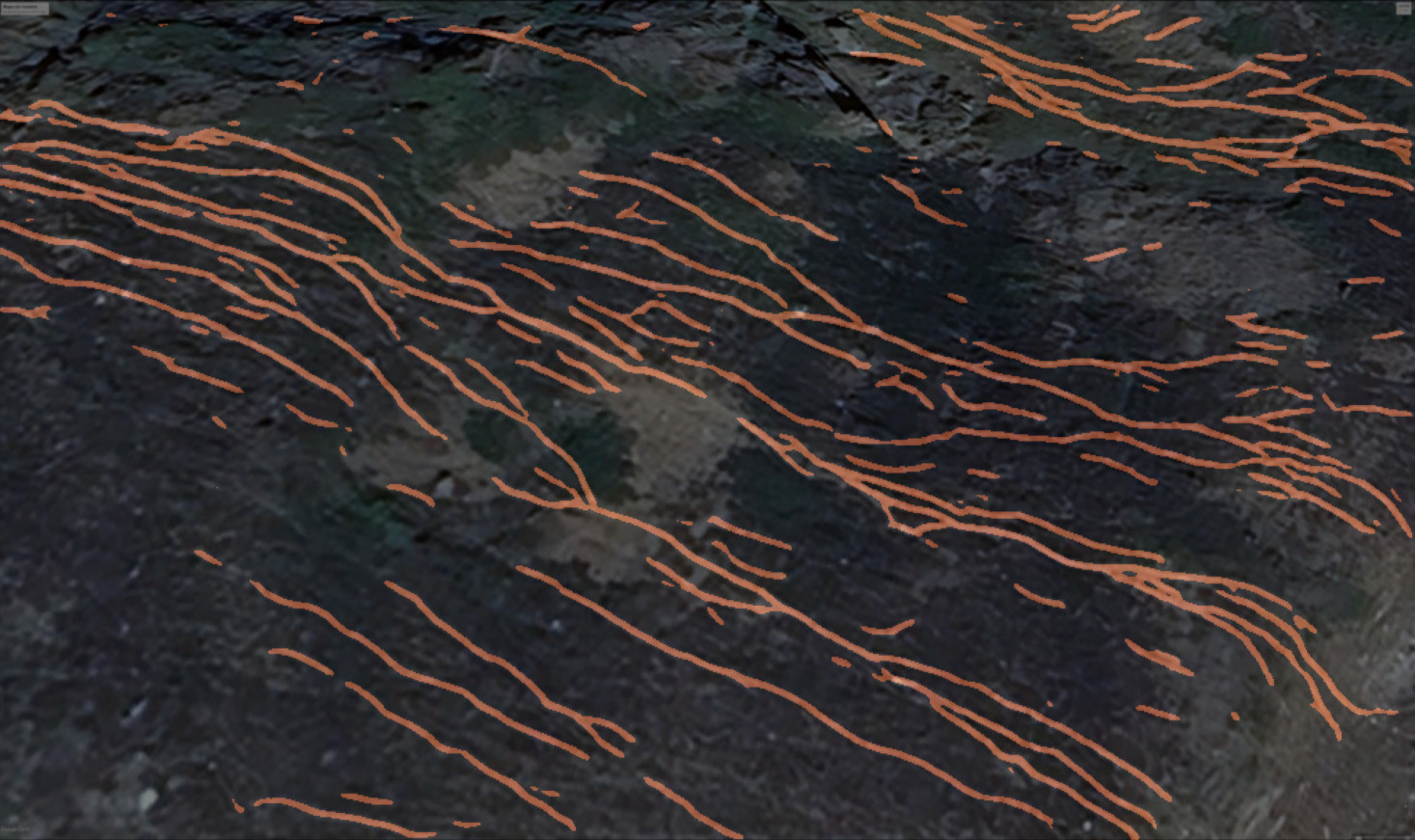}}

    \subfigure{\includegraphics[width=.48\textwidth]{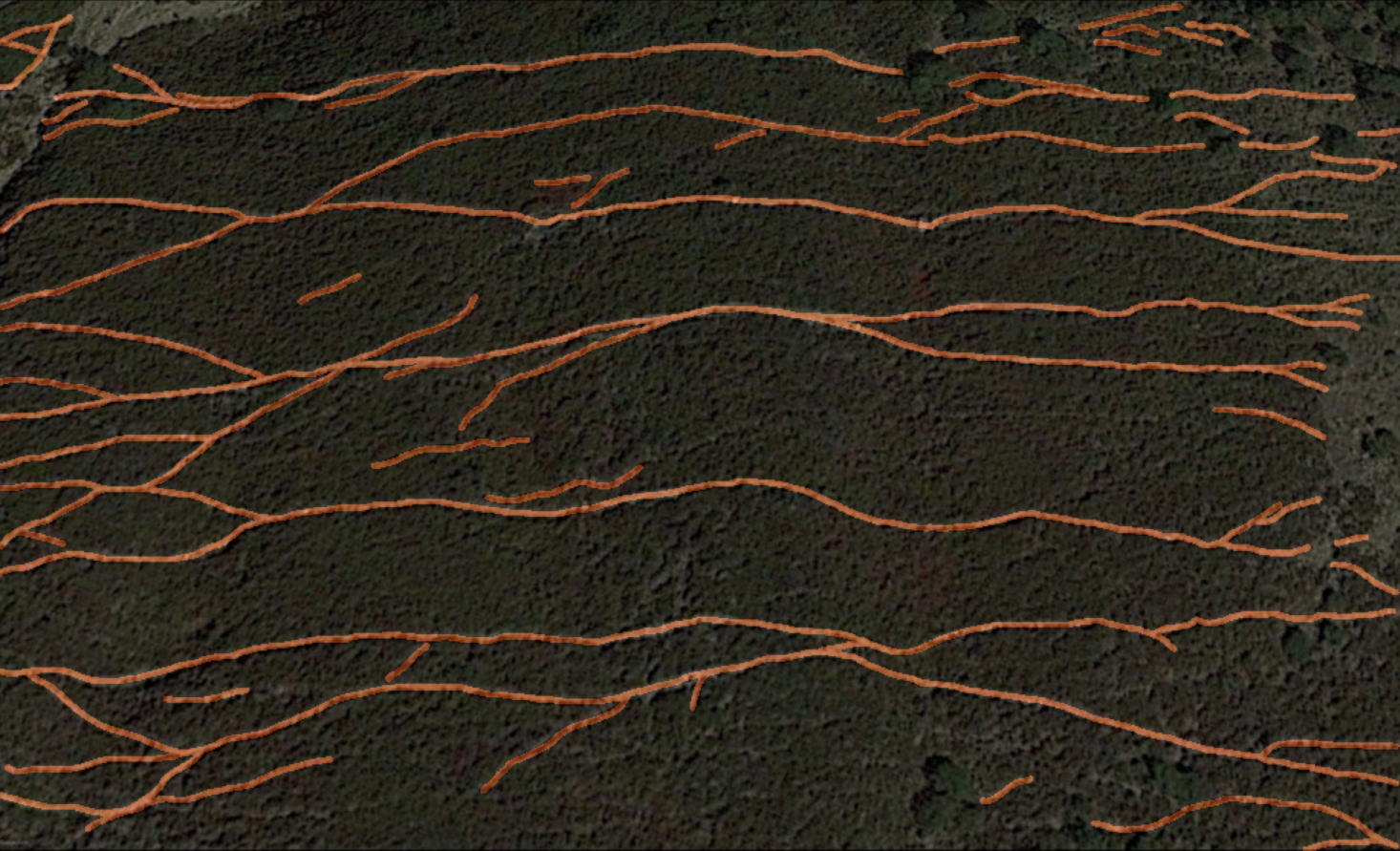}}~
    \subfigure{\includegraphics[width=.48\textwidth]{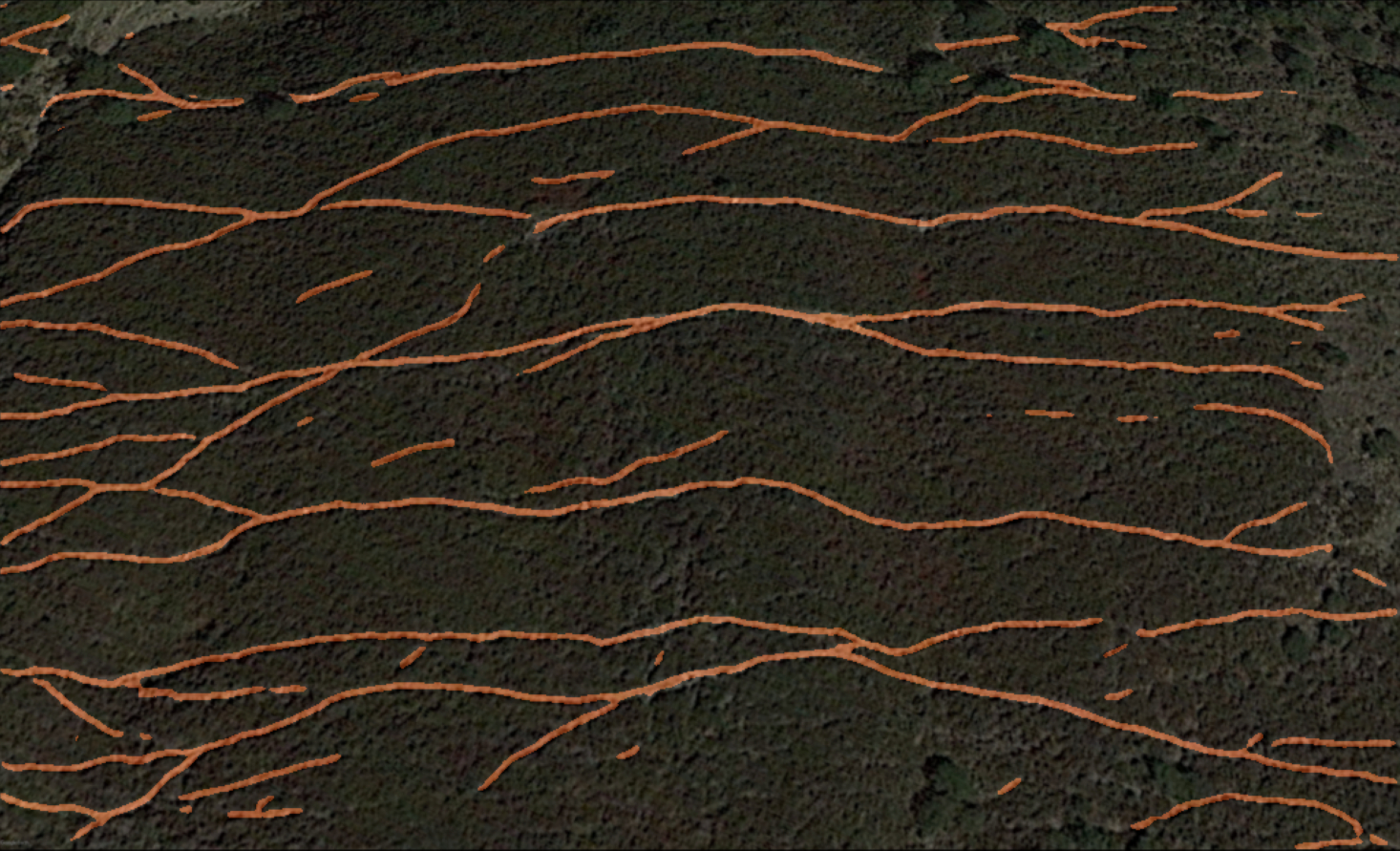}}

    \subfigure{\includegraphics[width=.48\textwidth]{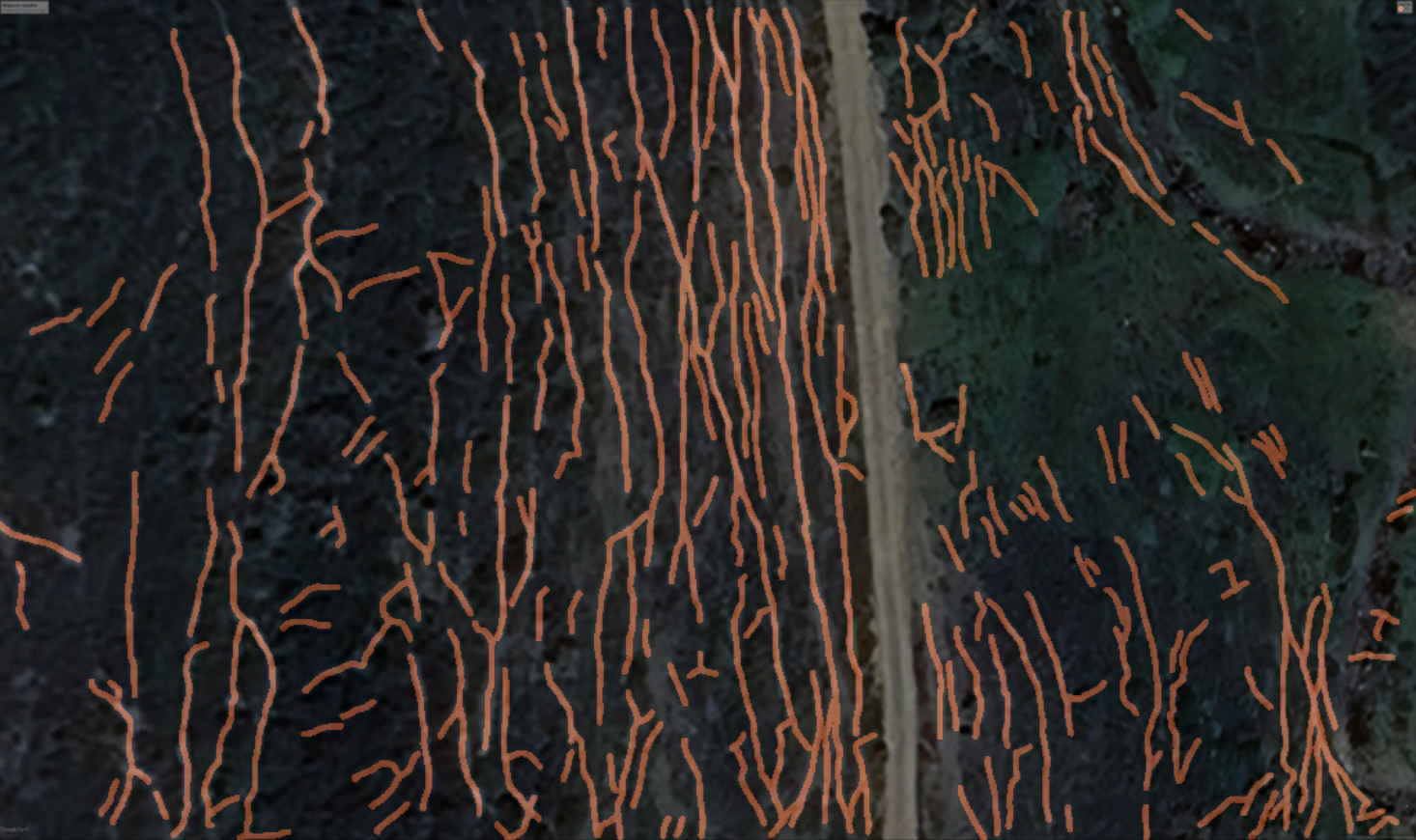}}~
    \subfigure{\includegraphics[width=.48\textwidth]{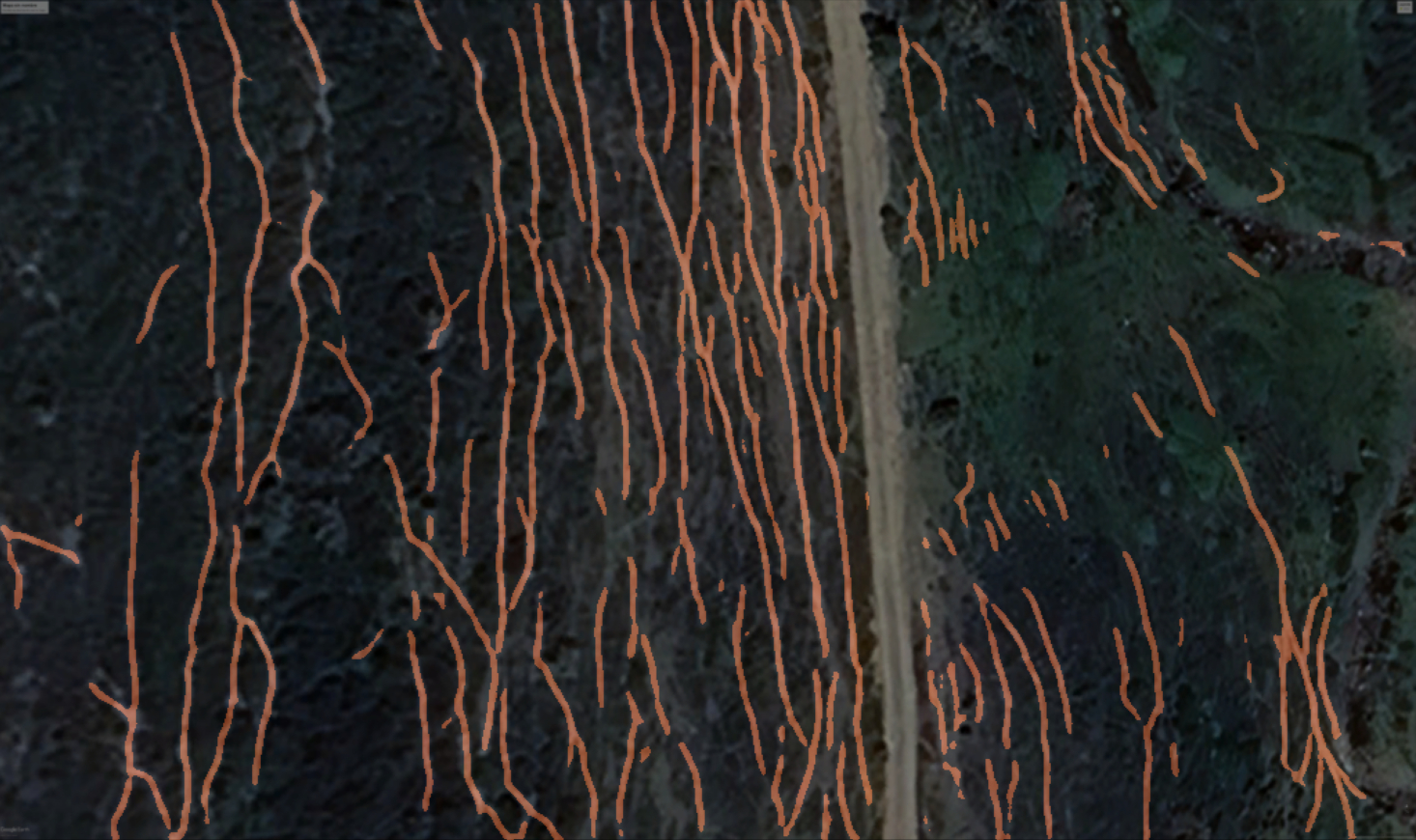}}

    \subfigure{\includegraphics[width=.48\textwidth]{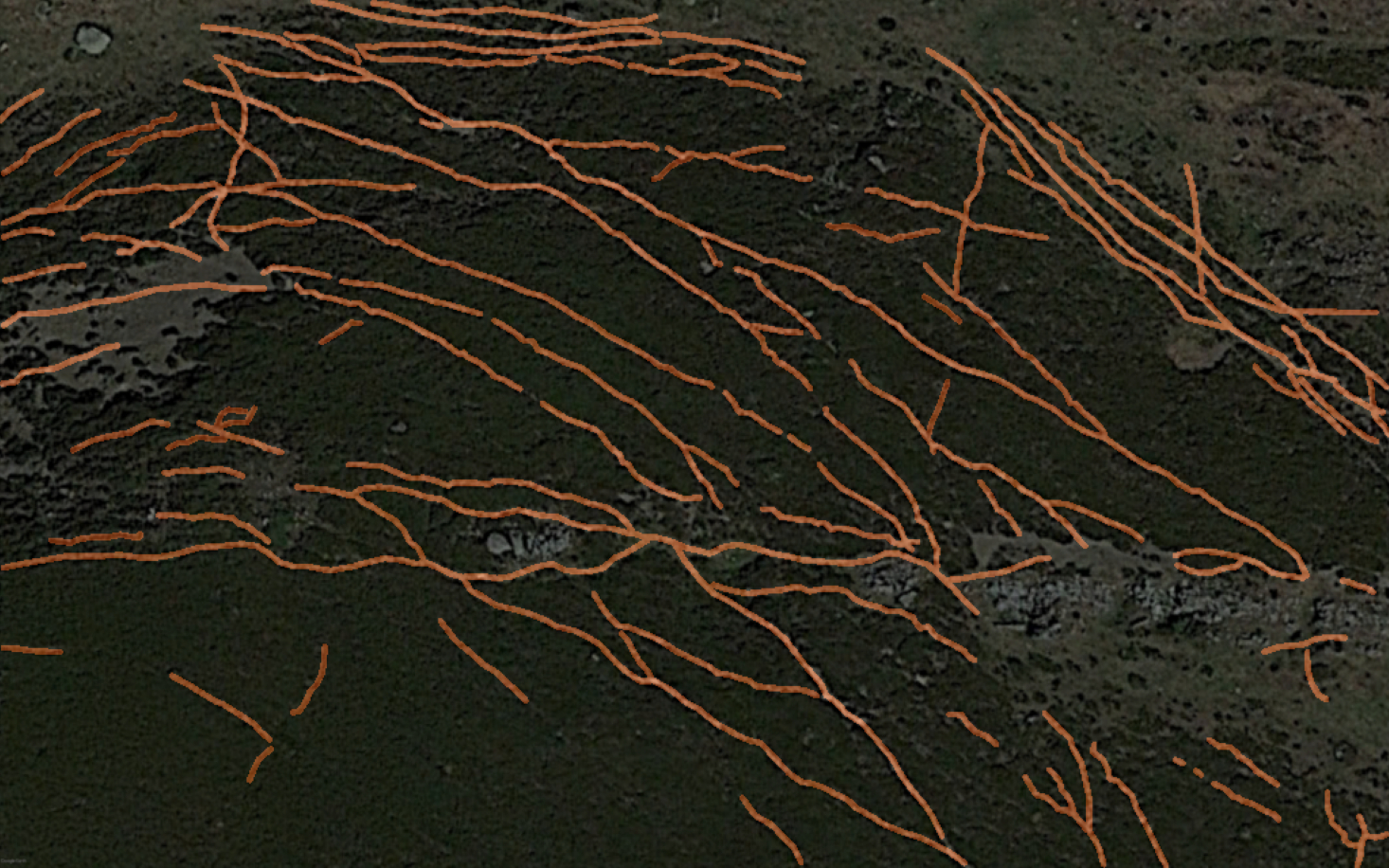}}~
    \subfigure{\includegraphics[width=.48\textwidth]{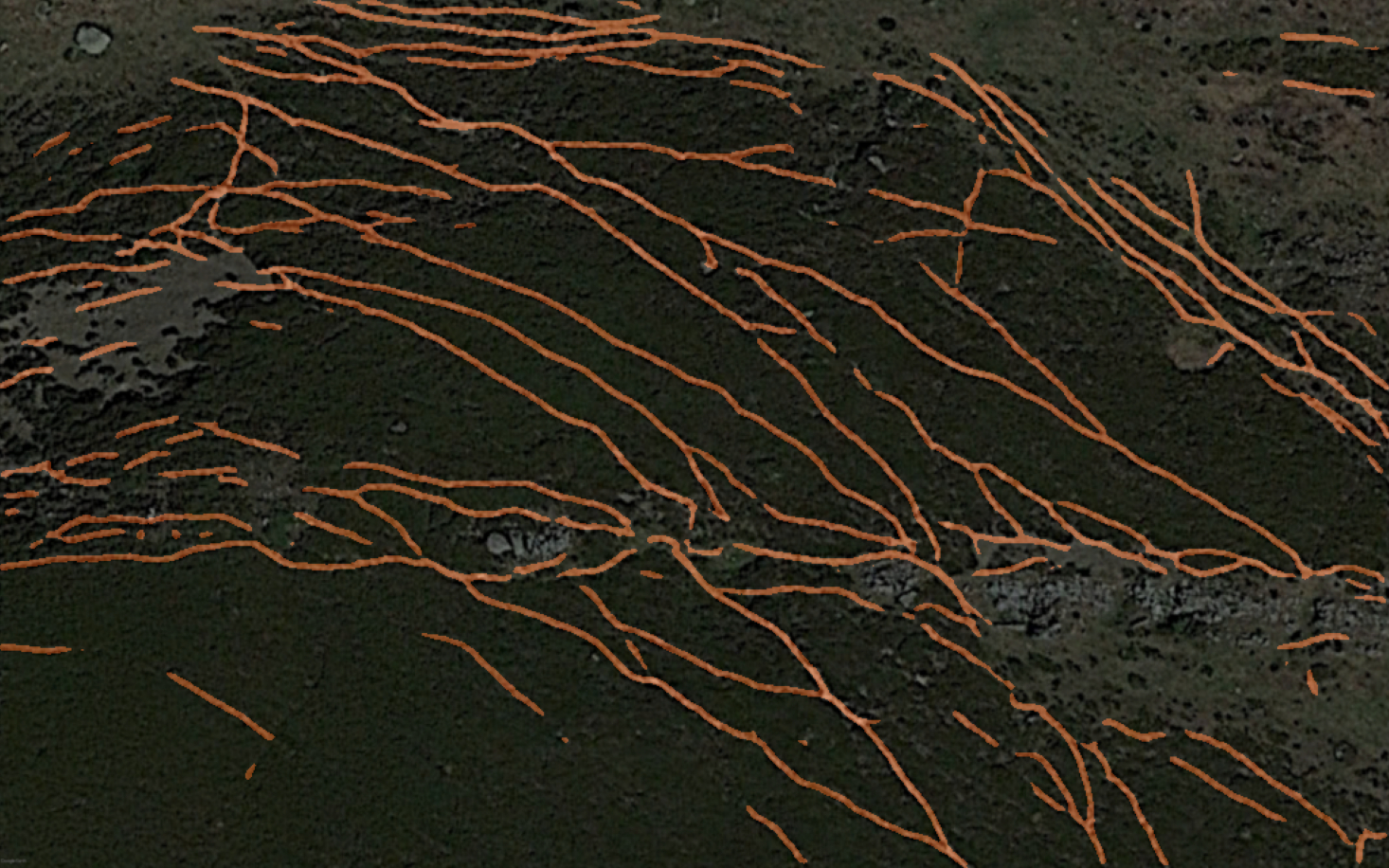}}
    \caption{Example of the original RGB \textit{groundtruth} image (left columns) and the result of the automatic trail segmentation (right column).
    The trails are highlighted in orange. Best viewed in colour.}
    \label{fig:imgs_gt_unet}
\end{figure*}

Our analysis indicates that the proposed semantic segmentation approach effectively classified herbivore trails by accurately predicting the probability that a pixel is part of a grazing trail with high precision. Figure~\ref{fig:imgs_gt_unet} presents examples of segmentation results produced by the optimal encoder-architecture combination (right), alongside their corresponding \textit{groundtruth} images (left).

It should be noted that semantic segmentation predictions effectively differentiate between grazing trails and anthropogenic structures, such as motorized dirt tracks. For instance, in Figure~\ref{fig:imgs_gt_unet}, the dirt track and its ditches were not misclassified as grazing trails, indicating the efficacy of semantic segmentation and its ability to: \textit{(i)} utilize non-local contextual information effectively and \textit{(ii)} maintain scale invariance in trail characterization.

This discriminative capability extends consistently to other landscape features in our dataset, including: 2oodland canopy edges, rock formations, and artificial structures, among others.
These results show significant promise for implementing such models in automated detection tools within geographic information systems, particularly for ecological monitoring and landscape management applications.

Given that five architectures were tested in combination with 14 encoders, the results exhibit considerable variability.
Thus, it is essential to consider both the architecture and the encoder, as some encoders outperform others.

The UNet architecture with the MambaOut encoder achieved the highest overall performance across combinations of architectures and encoders, with UNet and MIT B5 encoder yielding a similar performance.
Overall, the findings are consistent across the two metrics (IoU and $F_1$ score), used to compare the performance of the methods, with UNet achieving the best performance, and PSPNet the least.
Regarding encoders, MIT B5, MambaOut, and ConvNeXt exhibit a superior performance than the rest.

\begin{figure*}
    \centering
    \includegraphics[width=\linewidth]{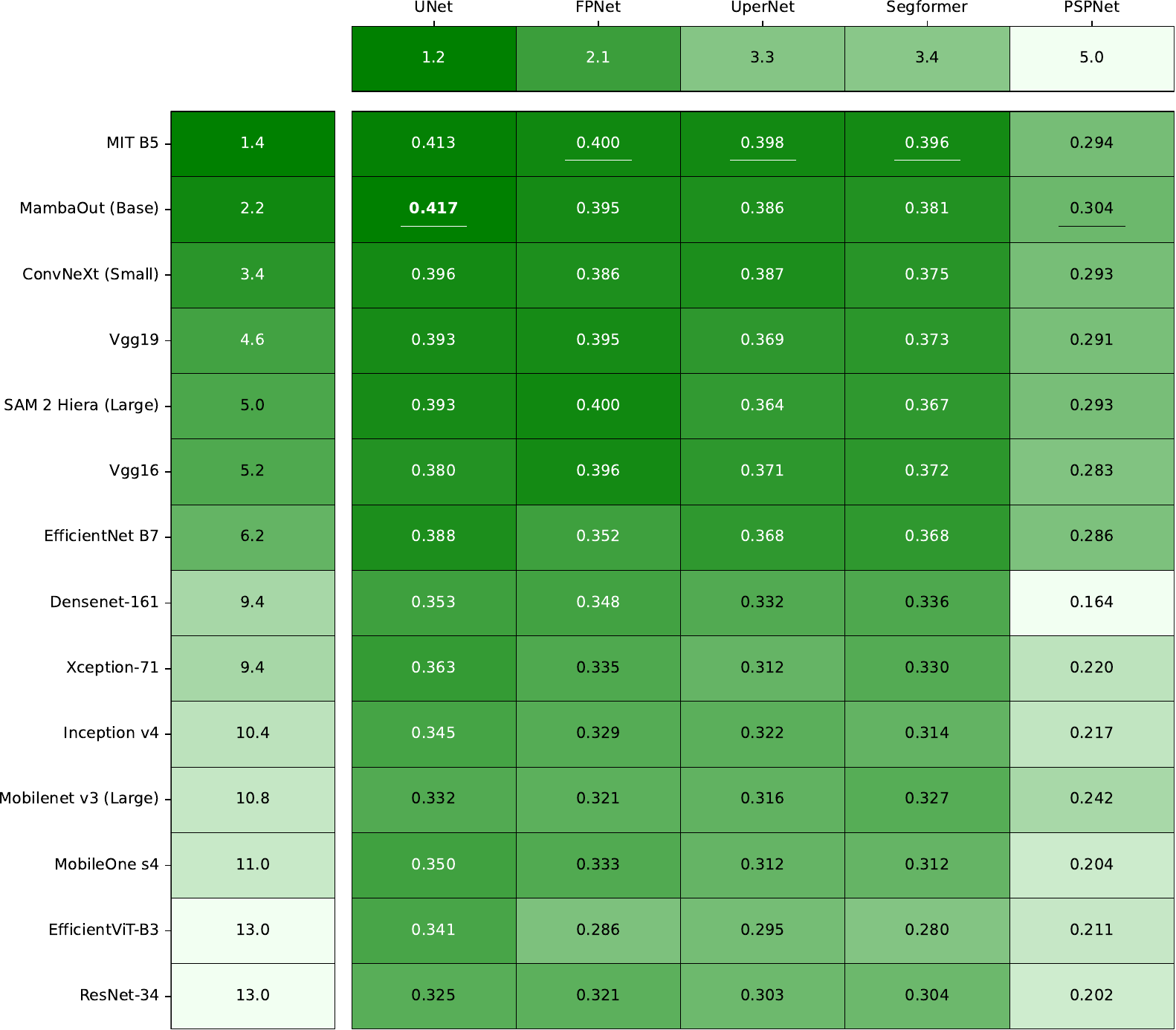}
    \caption{Heatmap of Intersection over Union (IoU) performance across all encoder-architecture combinations. Each cell displays the mean IoU for a specific pair, with underlined values indicating the best encoder for each architecture and bold values showing the overall top combination. The row above the heatmap matrix presents architecture rankings (averaged across encoders), while the column at the left shows encoder rankings (averaged across architectures), where lower values indicate better performance. Throughout the figure, darker green shading consistently represents better performance: in the main matrix this corresponds to higher (better) IoU values, while in the rankings it indicates lower (better) average rank values.}\label{fig:heatmap_iou}
\end{figure*}

Figure~\ref{fig:heatmap_iou} presents the Intersection over Union (IoU) performance metrics across all encoder-architecture combinations.
The visualization consists of:

\begin{enumerate}
\item Main heatmap matrix:
  \begin{itemize}
  \item Each cell displays the mean IoU score for an encoder-architecture pair.
  \item Colour intensity (green gradient) represents performance level (the darker the higher IoU).
  \end{itemize}
\item Performance Rankings:
  \begin{itemize}
  \item \emph{Row above matrix}: Average architecture ranking across encoders.
  \item \emph{Left column}: Average encoder ranking across architectures.
  \item Ranking convention: Lower values indicate better performance (with corresponding darker green shading).
  \end{itemize}
\item Top Performers Identification:
  \begin{itemize}
  \item Underlined values: Best encoder for each architecture.
  \item Bold value: Overall best-performing encoder-architecture combination.
  \end{itemize}
\end{enumerate}

The heatmap provides immediate visual comparison of segmentation performance, with colour intensity consistently representing superior performance (higher IoU or lower ranking values) through darker green
shading.

\begin{figure*}
    \centering
    \includegraphics[width=\linewidth]{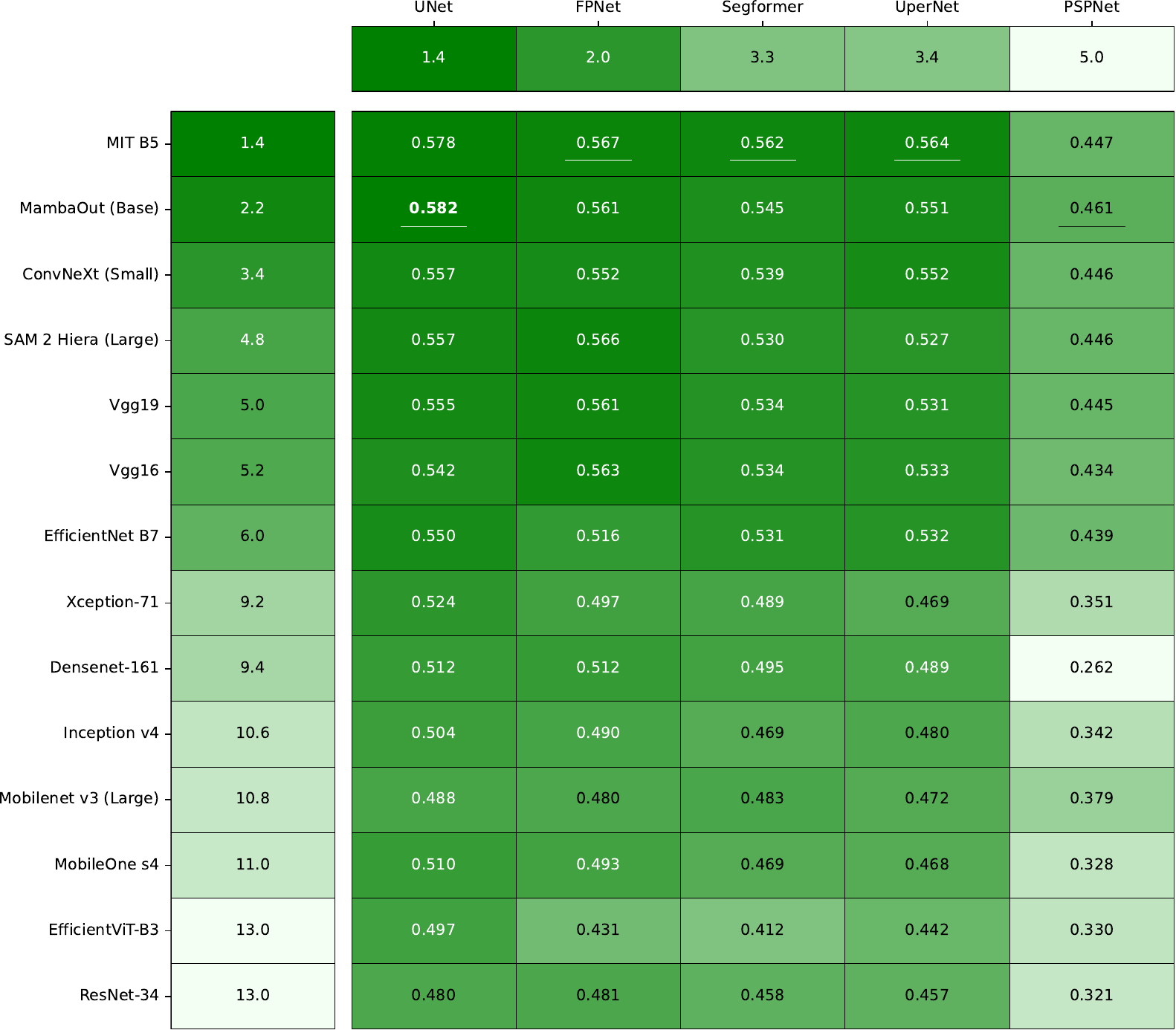}
    \caption{Heatmap of the $F_{1}$ performance measure across all encoder-architecture combinations. Each cell displays the mean IoU for a specific pair, with underlined values indicating the best encoder for each architecture and bold values showing the overall top combination. The row above the heatmap matrix presents architecture rankings (averaged across encoders), while the column at the left shows encoder rankings (averaged across architectures), where lower values indicate better performance. Throughout the figure, darker green shading consistently represents better performance: in the main matrix this corresponds to higher (better) $F_1$ values, while in the rankings it indicates lower (better) average rank values.}\label{fig:heatmap_f1}
\end{figure*}

Figure~\ref{fig:heatmap_f1} follows the same structure  but presents the results for the $F_{1}$ score.
The findings are consistent across both metrics,
with UNet being the top-performing architecture, followed by FPNet, while PSPNet consistently yields the worst performance.
Among encoders, MIT~B5, MambaOut, and ConvNeXt give the best results.
Notably, the UNet-MambaOut combination achieves the highest overall performance, though UNet paired with MIT~B5 produces nearly comparable results, suggesting both encoders are highly effective choices for UNet architecture.

With the aim to assess whether the differences among the architectures are significant or not, Bayesian test were used.
Since Bayesian test can only perform pair comparisons, the best architecture-encoder pairs were selected.
Overall, UNet performed significantly better than the rest of the architectures.
The differences between FPNet, UperNet, and Segformer were not statistically significant.
While PSPNet performed significantly worse than the rest.

\begin{figure*}
    \centering
    \includegraphics[width=.8\linewidth]{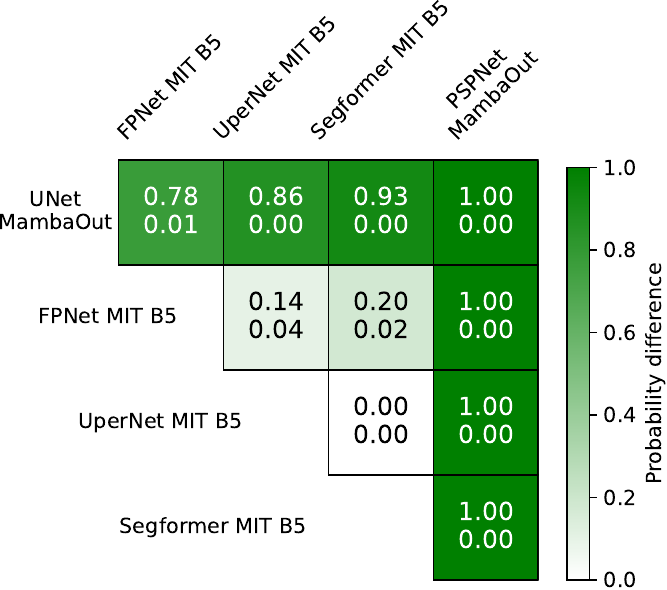}
    \caption{Bayesian tests heatmap, comparing the performance of the five architectures with their best encoder. Each cell contains the probability that the method in the row is better than the method in the column ($p_{left}$) on top and the probability that the opposite occurs ($p_{right}$) on the bottom. The green colour \textit{intensity} associated with each cell represents the $(p_{\text{left}} - p_{\text{right}})$ value.}\label{fig:BayesTests}
\end{figure*}

Figure~\ref{fig:BayesTests} presents the results of the Bayesian tests.
Each cell shows the $p_{\text{left}}$ (top) and $p_{\text{right}}$ (bottom) values.
The colour associated with each cell represents/\textit{codifies} the value: $(p_{\text{left}} - p_{\text{right}})$, \textit{i.e}., the probability that the method in the row is better than the method in the column minus the probability that the opposite occurs.
The main diagonal has not been represented since it is always $0$, \textit{i.e.}, a method is compared against itself.

\section{Discussion} \label{dis}

Our study shows that semantic segmentation is a powerful tool that can accurately identify grazing trails produced by large herbivorous mammals, and this was based on the capability of semantic segmentation for pixel classification, an advantage against, other approaches used in previous studies~\citep{hellman_detection_2020}.


Some approaches in automatic characterization have included the detection of continuous trampling activity of large herbivores based, among other factors, on the slope angle and relief~\citep{Godone_et_al_2018}, abrupt elevation changes~\citep{DIAZVARELA2014117}, topographic position index~\citep{Sas2012DETECTIONOO}, differentiation of terraces from natural plains using correlation~\citep{SOFIA2014123}, line segment detection methods applied to terrace walls~\citep{Bailly_Lavavaseur_2012}, the use of digital elevation models~\citep{PIJL2020101977}, and the analysis of high-resolution images and classification based on the Fourier transformation~\citep{hellman_detection_2020}.


The morphology of the grazing trails network in our images mainly featured parallel paths over branched or radial ones. However, our models were able to detect these other morphologies, making them useful for applications in other \textit{movement} scenarios, such as spatial displacement toward focal resources like water puddles in the savannah, natural salt licks, and grazing-rich spots, to consider a few examples.
It should be particularly stressed that (as previously stated) semantic segmentation predictions effectively differentiate between grazing trails and anthropogenic structures, such as motorized dirt tracks.
This is an important feature since it may appear often in remote sensing imagery.


An important advantage of the approach we considered, when compared to that shown by \cite{hellman_detection_2020}, is that our method classifies grazing trails at pixel level, whereas \cite{hellman_detection_2020} classify a complete image window as a window where a grazing trail appears.
In particular, \cite{hellman_detection_2020} detect areas with the presence of grazing trails (using public accessible images from Google or Bing) and then created windows of $250\times250$ pixels in size ($37\times37$m). For each one of them, they inferred the so-called Fourier domain periodogram.
Periodograms display the squared magnitude of the discrete Fourier transform (DFT) (the so-called power spectrum) at all spatial frequencies of the image.
From the selected image, and for each window, a $95^{th}$ percentile histogram thresholding technique is applied, and a Fourier band-pass post-processing operation, afterwards.
The inverse DFT completes the processing step. This resulting enhanced image is partitioned into patches, and each patch is then classified as grazing trail or not. However, this discretization on a window level fails to characterize grazing trails as a continuous and quantified feature and depending on the environment they can be more or less detectable.

Our approach supersedes \cite{hellman_detection_2020} methodology as we are able to identify pixels that composed the trails  (\textit{i.e}., classification at a pixel level).
This facilitates comparing different images of the same area over time and applying topology and spatial analysis to understand the morphological changes, reduction or increase in trails density, and their relation to changes in the activity of large herbivores. All these features are considerably useful in landscape monitoring programs.

Grazing trails develop over heterogeneous land cover and topography that produces a gradient in the visibility of the trails, ranging from the highly distinguishable to the barely discernible~\citep{hellman_detection_2020}.
This leads to unavoidable errors, inherent to observer subjectivity when labelling these features~\citep{powell_sources_2004} and our work is not free of this issue. In order to minimise subjectivity error in the labelling of trail images (\textit{i.e}., when a fading trails begins or ends) it is advisable to use high-resolution images and to define a protocol describing the threshold to decide when a trail exists or not, which could be based on a trail continuity distance parameter~\cite{hellman_detection_2020}.
Our proposal is to use these methodologies for the automatic monitoring of grazing trails, allowing for their identification, quantification, and tracking over time, for example within Geographic Information System (GIS) tools.

For the semantic segmentation algorithms to be effectively used in GIS tools it is essential that these algorithms (i) minimise type 1 errors, meaning they should not mistakenly identify other landscape features or infrastructures as grazing trails (\textit{e.g}., structures with edges such as tree patches, dirt roads, fences, walls, agricultural furrows, etc.), and (ii) ensure they do not underestimate the amount of grazing trails present. The first point is particularly crucial, as the second would still allow for the quantification and monitoring of changes in grazing trails.
Our results suggest exactly that: while some grazing trail segments are not identified, other non-grazing trail structures have not been incorrectly classified as grazing trails.
A clear example of this can be seen in Figure~\ref{fig:imgs_gt_unet}, where the edges of a dirt road have not been classified as grazing trails.
Despite these caveats, our methodology has proven  to be useful to monitor grazing trails in a variety of habitats and shapes, which can be used to monitoring trail changes over time and its potential impact on habitats as a proxy of herbivory intensity.



\section{Conclusions and future work} \label{concluFW}

This paper presents the first pixel-level semantic segmentation approach for mapping grazing trails created by large herbivores, advancing beyond the limitations of current patch-based detection methods. Whereas existing approaches can only determine the presence / absence of trails within coarse $37\times 37m$ patches~\citep{hellman_detection_2020}, our method achieves precise trail detection at the original pixel resolution of the image. This could be used to reconstruct the full extent of the grazing trails, enabling precise measurement of trail morphology, connectivity, and spatial distribution patterns.

Through the systematic evaluation of several modern semantic segmentation architectures and encoding methods, it was determined that the best model architecture is UNet, closely followed by FPNet, UperNet, and Segformer. In contrast, PSPNet showed the worst performance. Based on these findings, we recommend UNet for future studies in this domain and advise against the use of PSPNet. Additionally, another interesting result of this study is the important influence of the encoding method on model performance, with MIT B5 and MambaOut (Base) performing the best across all the architectures tested.

One of the most interesting potential uses of the proposed semantic segmentation algorithms is the mapping and analysis of the temporal variation of grazing trails to assess changes in herbivory intensity. On the other hand, it would be interesting to test the efficiency of the methods used in our study in other areas and grazed ecosystems. Future work would include the integration of these methods into geographic information system tools that would facilitate their use by land-managers and stakeholders to assist in landscape management and conservation programs.



\newpage

\section*{Acknowledgments}
This work is partially supported by Asturias Biodiversity Complementary Program BIO06 (Next Generation EU/PRTR), Strategic Projects Oriented Towards Ecological and Digital Transition (TED2021-131388B-100), Spanish Knowledge Generation Projects (PID2023-146074OB-I00) funded by EU Next Generation and Spanish Research Agency, and Spanish Research Council Tenured Scientist Incorporation Grants 2022 (202230I041).
J.F.~D\'iez-Pastor was supported by a mobility grant PRX22/00634 from the Spanish Ministry of Universities.
Some images of the graphical abstract have been designed with \texttt{Flaticon.com} resources.

\section*{Computer Code Availability}
Datasets will be accessible via the HuggingFace Hub (\url{https://huggingface.co/datasets/jfdpastor/Comp_Geo_Herb_Trails_8x}).
Raw Images and code will be accessible in GitHub. (\url{https://github.com/joseFranciscoDiez/HerbiSeg}).

\section*{Credit authorship contribution statement}
\textbf{José Francisco Díez-Pastor:} Conceptualization, Formal Analysis, Funding acquisition, Methodology, Software,  Supervision, Validation, Visualization, Writing -- original draft, Writing -- review \& editing.

\textbf{Francisco~Javier Gonz\'alez-Moya:} Software, Validation, Visualization, Writing -- original draft, Writing -- review \& editing

\textbf{Pedro Latorre-Carmona:} Conceptualization, Formal Analysis, Funding acquisition, Investigation, Methodology,  Project administration, Resources, Software, Supervision, Validation, Visualization, Writing -- original draft, Writing -- review \& editing.

\textbf{Francisco Javier Pérez-Barbería:} Ecological conceptualization, Data collection, Funding acquisition, Supervision, Writing -- original draft, Writing -- review \& editing.

\textbf{Ludmila I. Kuncheva:} Formal analysis, Writing -- original draft, Writing -- review \& editing.

\textbf{Antonio Canepa-Oneto:} Conceptualization,  Writing -- original draft, Writing -- review \& editing.

\textbf{\'Alvar Arnaiz-Gonz\'alez:} Conceptualization, Methodology,  Supervision, Writing -- original draft, Writing -- review \& editing.

\textbf{César García-Osorio:} Formal analysis, Funding acquisition, Writing -- review \& editing.

\section*{Declaration of competing interest}
Authors declare that they have no known competing financial interests or personal relationships that could have appeared to influence the work reported in this paper.

\bibliographystyle{cas-model2-names}
\bibliography{HerbivRefs}

\end{document}